%% file: main.tex
\documentclass[conference]{IEEEtran}
\IEEEoverridecommandlockouts
\def\BibTeX{{\rm B\kern-.05em{\sc i\kern-.025em b}\kern-.08em
    T\kern-.1667em\lower.7ex\hbox{E}\kern-.125emX}}
\usepackage[color,matrix,arrow,all]{xy}
\usepackage{multirow}
\usepackage{float}
\usepackage{subfigure}
\usepackage{graphicx}
\usepackage{subfigure}
\usepackage{balance}  
\usepackage{amsmath,amssymb,amsfonts}
\usepackage{enumitem}
\usepackage{booktabs}

\usepackage{framed}
\usepackage{color}
\usepackage{cuted}

\usepackage[bookmarks=false]{hyperref}
\usepackage{balance} 
\usepackage{listings,upquote,soul,framed}

\usepackage[dvipsnames]{xcolor}  
\usepackage{listings}  
\usepackage{lineno} 

\usepackage{algorithm}
\usepackage{algorithmicx} 
\usepackage{algpseudocode} 

\usepackage{bbding}
\usepackage{setspace}
\usepackage{array}
\usepackage{hyperref}
\usepackage{url}
\usepackage{cleveref}
\hypersetup{colorlinks,
    linkcolor=black,
    filecolor=black,      
    urlcolor=blue,
    citecolor=black,
    pdfstartview=Fit,
    breaklinks=true
}

\makeatletter
\newif\if@restonecol
\makeatother

\makeatletter
\newcommand{\removelatexerror}{\let\@latex@error\@gobble}
\makeatother
\allowdisplaybreaks[4]

\newcommand{\lee}[1]{\textcolor{black}{#1}}

\AtBeginDocument{%
  \providecommand\BibTeX{{%
    \normalfont B\kern-0.5em{\scshape i\kern-0.25em b}\kern-0.8em\TeX}}}

\begin{document}





\title{Effective Few-Shot Named Entity Linking by Meta-Learning}

\setcounter{figure}{0}

\author{
	\IEEEauthorblockN{
	Xiuxing Li \IEEEauthorrefmark{1}\IEEEauthorrefmark{4}, Zhenyu Li\IEEEauthorrefmark{1}\IEEEauthorrefmark{4}\thanks{\IEEEauthorrefmark{4} Xiuxing Li and Zhenyu Li contribute equally.},
	Zhengyan Zhang\IEEEauthorrefmark{1},
     Ning Liu\IEEEauthorrefmark{3},
     Haitao Yuan\IEEEauthorrefmark{1},
     Wei Zhang\IEEEauthorrefmark{2}, \\
     Zhiyuan Liu\IEEEauthorrefmark{1},
     Jianyong Wang\IEEEauthorrefmark{1}\IEEEauthorrefmark{10}\thanks{\IEEEauthorrefmark{10} Corresponding author.} 
     }
	\IEEEauthorblockA{\IEEEauthorrefmark{1}Department of Computer Science and Technology, Tsinghua University, Beijing, China\\
		Email: \{lixx16,zhengyanzhang14,yht16\}@mails.tsinghua.edu.cn,
		lizhenyucs@gmail.com,\\
		liuzy@tsinghua.edu.cn,
		jianyong@mail.tsinghua.edu.cn
		}

	\IEEEauthorblockA{\IEEEauthorrefmark{2}School of Computer Science and Techniques, East China Normal University, Shanghai, China\\
		Email: zhangwei.thu2011@gmail.com
}
	
	\IEEEauthorblockA{\IEEEauthorrefmark{3}School of Software, Shandong University, Jinan, China\\
		Email: victorliucs@gmail.com}
	}

\maketitle
\pagestyle{empty}

\input{sec0-abstract}
\input{sec1-introduction}

\input{sec3-preliminary}
\input{sec4-framework}

\input{sec4-methodology}

\input{sec5-implementation}
\input{sec6-evaluation}
\input{sec2-background_and_overview}
\input{sec7-future_work}
\input{sec9-conclusion}
\input{sec-ack}

\bibliographystyle{IEEEtran}  
\bibliography{main}
\clearpage


\end{document}

%% file: sec0-abstract.tex
\begin{abstract}
Entity linking aims to link ambiguous mentions to their corresponding entities in a knowledge base, which is significant and fundamental for various downstream applications,  e.g., knowledge base completion, question answering, and information extraction. While great efforts have been devoted to this task, most of these studies follow the assumption that large-scale labeled data is available. However, when the labeled data is insufficient for specific domains due to labor-intensive annotation work, the performance of existing algorithms will suffer an intolerable decline. 
In this paper, we endeavor to solve the problem of few-shot entity linking, which only requires a minimal amount of in-domain labeled data and is more practical in real situations.
Specifically, we firstly propose a novel weak supervision strategy to generate non-trivial synthetic entity-mention pairs based on mention rewriting.
Since the quality of the synthetic data has a critical impact on effective model training, we further design a meta-learning mechanism to assign different weights to each synthetic entity-mention pair automatically.
Through this way, we can profoundly exploit rich and precious semantic information to derive a well-trained entity linking model under the few-shot setting.
The experiments on real-world datasets show that the proposed method can extensively improve the state-of-the-art few-shot entity linking model and achieve impressive performance when only a small amount of labeled data is available. Moreover, we also demonstrate the outstanding ability of the model's transferability. Our code and models will be open-sourced. 
\end{abstract}

%% file: sec1-introduction.tex
\section{Introduction}\label{sec-intro}

Knowledge bases, such as DBpedia~\cite{Auer2007DBpedia}, YAGO~\cite{fabian2007yago}, Freebase~\cite{bollacker2008freebase} and WordNet \cite{miller1998wordnet}, have been proposed to manage enormous amount of information due to the exponentially increasing of data in recent years~\cite{survey_vldbj,deepeye_icde}. In particular, knowledge bases contain extensive information about the real worlds' entities, semantic classes, and mutual relationship, which are mostly extracted from the data in the form of natural language. However, the same entity may correspond to different names/words/mentions, and different entities may mean the same name/word/mention in different corpora/domains/sentences, which would cause the ambiguities or errors in knowledge bases. Hence, entity linking, aiming to link ambiguous mentions to their corresponding entities, has drawn extensive attention from both academia and industry, and massive prominent methods have been proposed~\cite{mann2003unsupervised, artiles2007semeval,yin2007object, han2009named, hachey2013evaluating, nvbench}. In addition, entity linking can facilitate various tasks such as knowledge base completion~\cite{dong2014knowledge, lehmann2015dbpedia,shi2018open}, question answering~\cite{diefenbach2018core, yin2016simple, veyseh2016cross}, and information extraction~\cite{hasibi2016exploiting, corcoglioniti2016knowledge, hasibi2015entity}. For example, many question answering systems, benefiting from the development of entity linking technology, can answer users' questions more precisely; entity linking can also improve the accuracy of information extraction results due to the consideration of the semantic context. In summary, entity linking is an essential technique for knowledge bases since it resolves the lexical ambiguity of entity mentions and determines their meanings.

The essential challenge of entity linking task is how to embed the mention context and entities more accurately. The early methods based on statistics mainly utilized feature engineering to construct semantic representation~\cite{he2013learning},~\cite{newman2018jointly},~\cite{radhakrishnan2018elden,deepeye_tkde}. Inspired by the rapid development of deep learning technology, a new generation of approaches for entity linking based on neural models and deep learning has emerged. They have managed to excel in entity linking as in many other natural language processing tasks due to their ability to learn useful deep distributed representations of linguistic data~\cite{JunyoungChung2014Empirical},~\cite{young2018recent,ncnet}, and significant improvements over existing approaches~\cite{chisholm-hachey-2015-entity},~\cite{kraljevic2021multi},~\cite{ratinov2011local} have been observed.

\begin{figure}
  \centering
  \includegraphics[width=0.8\linewidth]{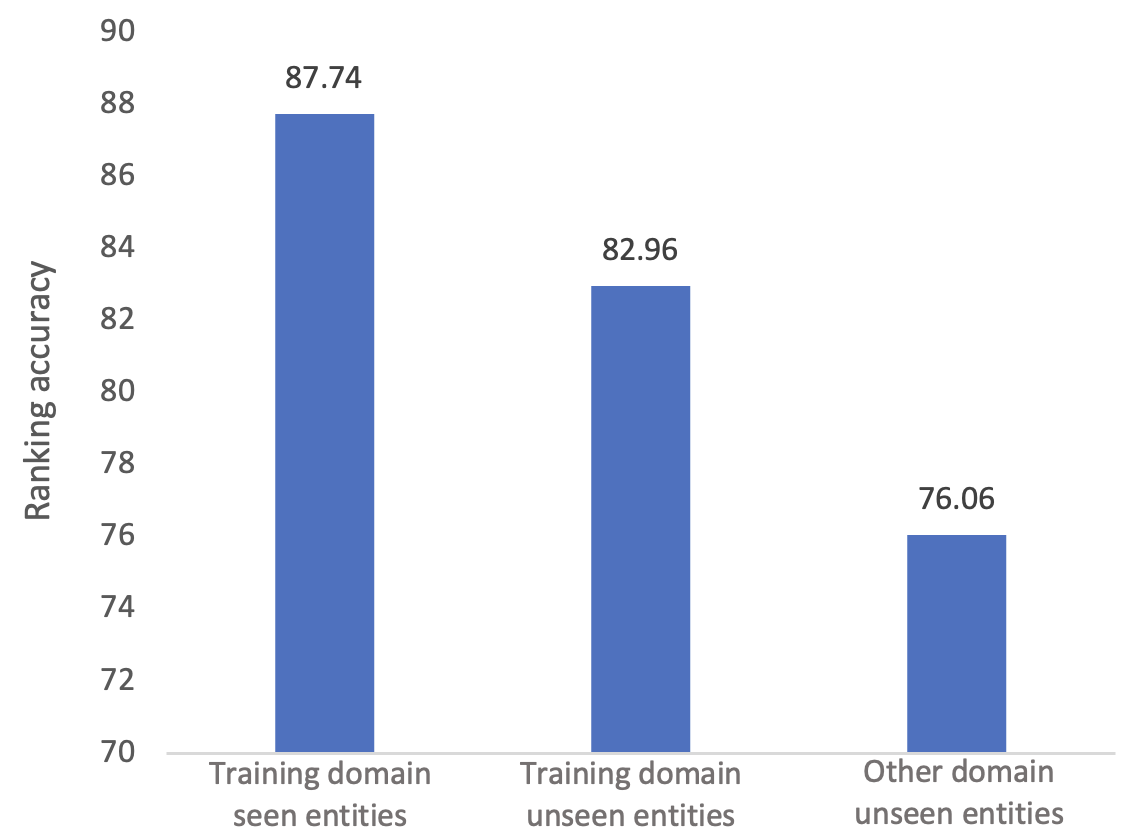}
  \caption{The performance of existing models degrades dramatically when training data of the corresponding entities and the target domain is insufficient.}
  \label{performance when lacking domain knowledge}
\end{figure}

 Nevertheless, these methods are still not adequate for real-world entity linking problems. This is because they generally require a large amount of annotation data as the prerequisite. In the absence of sufficient labeled data, the learning ability of these methods will rapidly decline, which significantly restricts the applicability of these methods. Unfortunately, the number of announced data remains far from complete concerning the vast number of real-world entities, as the statistics of each domain shown in Table~\ref{Zero-Shot Entity Linking Dataset}.  Logeswaran et al.~\cite{DBLP:conf/acl/LogeswaranCLTDL19} shows that the linking accuracy of a full-transformer model will encounter performance degradation when lacking training data of the corresponding entities in a target domain (see Figure~\ref{performance when lacking domain knowledge}). 
 On the other hand, as a main difficulty of entity linking is the lexical ambiguity of entities, most existing algorithms rely on powerful resources of the knowledge base~\cite{milne2008learning}, such as alias table~\cite{fevry2020empirical}, frequency statistics~\cite{spitkovsky2012cross}, entity type~\cite{cucerzan2007large}, etc. However, such powerful resources are not available in specific entity dictionaries (e.g., legal cases, company project descriptions). Most existing methods can't work well and their performance will suffer a severe decline, especially when predicting an entity that has not appeared before.

This makes the study of the few-shot entity linking problem gradually become a new trend~\cite{DBLP:conf/acl/LogeswaranCLTDL19},~\cite{gillick-etal-2019-learning},~\cite{wu-etal-2020-scalable}. Two different optimization directions have been proposed to alleviate the lack of labeled data, and remarkable results have been achieved: (1) External information enhanced methods take advantage of different sources of external information to improve the quality of data, including manually curated mention tables~\cite{ganea-hofmann-2017-deep}, incoming Wikipedia link popularity~\cite{eshel-etal-2017-named}, and Wikipedia entity categories~\cite{gillick-etal-2019-learning}. Among external information sources, Wikipedia has gradually gained widespread attention from academia. Though external source information can provide supplementary information, these methods inevitably have serious shortcomings. Firstly, introducing external information may introduce new errors; secondly, the algorithm's effectiveness is directly affected by the quality of the external data, and the applicability of the algorithm is inevitably restricted. 
(2) Pre-trained models, built on large-scale textual data, have involved much knowledge which might be useful for many relevant tasks.  Meanwhile, entity linking can obtain a better way to embed the context and entities by leveraging massive general domain knowledge from the pre-trained models like ELMo~\cite{peters-etal-2018-deep}, BERT~\cite{devlin-etal-2019-bert}, and their modifications. Recently, Wu et al. presents a two-stage BERT-based linking algorithm BLINK~\cite{wu-etal-2020-scalable} to solve the problem of insufficient data from the technical level by improving the representation ability of the model, which is the most relevant work to us and keeps the state-of-the-art performance in this branch. 

However, this direction is still in its infancy and has two main challenges that remained unresolved: (1) They directly use the pre-trained models to represent the context of the entities. Thus, the quality and quantity of knowledge that they introduced are completely determined by pre-trained models. In other words, the knowledge that they can acquire merely comes from the these models, which greatly limits the model's performance. (2) They cannot make adjustments according to the target domain to improve the quality of training data. This is because human language is astoundingly complex and diverse, and there exists a huge gap between different domains. 

To deal with the above issues, we present an effective few-shot entity linking system by leveraging a meta-learning mechanism (see Figure \ref{framework} for an overview). More specifically, we first adopt a weak supervision strategy to generate in-domain synthetic data. What's more, in the experiment process, we found that it is difficult to control the quality of synthetic data, especially in the specific domains. Therefore, we design a two-stage architecture to generate the synthetic data. Initially, we use a heuristic method to generate the mention-entity pairs. Then we adopt the weak supervision strategy to generate the synthetic data based on these mention-entity pairs. In this way, we can obtain high-quality synthetic data. Furthermore, we design a meta-learning mechanism to improve the quality of the synthetic data. Finally, we conduct extensive experiments on real-world datasets. As shown through detailed experiments, our meta-learning model-based approach achieves much better performance than the state-of-the-art methods on real-world datasets. 

The main contributions of this paper are summarized as follows:
\begin{itemize}
  \item We propose a weak supervision strategy to generate in-domain synthetic data to obtain more knowledge to deal with the ever more severe lack of labeled data. 
  
  \item Further, we design a  meta-learning mechanism to improve the quality of synthetic data by weighting it automatically.
  
  \item We propose a novel few-shot named entity linking system to break through the limitation of labeled data. 
  
  \item We conduct comprehensive experiments on real-world datasets with reasonable measurements, and the results show our model outperforms the state-of-the-art methods.
  
\end{itemize}

%% file: sec3-preliminary.tex
\section{Preliminary}\label{sec-preliminary}
In this section, we present some preliminary knowledge of the proposed model. Firstly, we list basic notations and definitions in Table~\ref{tab:tab-notations}. In the next, we define the relevant underlying elements of the few-shot entity linking problem. Afterward, we define the synthetic data generation process for weak supervision in detail. Finally, considering that meta-learning is an important architectural component of our system, we present a streamlined overview and analysis of the primary approaches to meta-learning.

\begin{table}[ht]
\setstretch{1.35}
    \centering
    \caption{Notations.}
    \begin{tabular}{|c|m{2in}|}
         \hline
       Notation  & Definition  \\
         \hline
        $G$ &  A knowledge base.   \\ 
        $E,e$ & Entity set of a knowledge base, and $e$ is an entity in knowledge base. \\ 
        $M,m$ & Mention set found in texts, and $m$ is a mention. \\ 
        $\phi^t$ & Model parameters at the $t$ th epoch. \\ 
        $w$ & The weight of samples in the training dataset. \\ 
        $T5(\tau(e)) $& T5 model with summary task. \\ 
        $L,l_i$ & Loss function, specifically, $l_i$ is the loss of $i$ th instance.  \\ 
         \hline
    \end{tabular}
    \label{tab:tab-notations}
\end{table}

\subsection{Entity Linking}
Before introducing the entity linking task, we first define some basic concepts.

\textbf{Entity.} \textit{An entity $e$ is a meaningful real-world object in a specific knowledge base. For example, an entity in Wikipedia can be described by the title and description of the Wikipedia page.  }

\textbf{Mention.} \textit{A mention is a textual phrase referring to some entity of a knowledge base such as Unified Medical Language System (UMLS)\footnote{https://www.nlm.nih.gov/research/umls/index.html},YAGO\footnote{https://yago-knowledge.org/} and so on.}

In free text, there are a tremendous amount of mentions which can provide useful information. Usually, the mentions in short texts are labeled manually or generated by named-entity recognition methods. Note that mentions may be represented by a short phrase or alias in the short texts. 

\textbf{Knowledge Base} \textit{A knowledge base is a machine-readable source that comprises entities and their relations. } We denote a knowledge base as $G = \{ E; R; T \}$, where $G$ is a directed graph in which nodes are entities and edges corresponding to the subject-property-object triples, $E$ indicates a set of entities, $R$ indicates a set of relations, and $T$ indicates a set of fact triples. Each triple (\textit{h}, \textit{r}, \textit{t}) $\in$ $T$ indicates there is a relation \textit{r} $\in$ $R$ between \textit{h} $\in$ $E$ and \textit{t} $\in$ $E$.

\lee{\textbf{Domain.} \textit{A domain means a specialized entity dictionary such as legal cases and company project descriptions. Each domain has many unique entities related to a specific theme. }}

\textbf{Entity Linking} \textit{Given an entity set denoted as $E$ from the knowledge base $G$ and a set of texts with mentions $M$ labeled, the goal of entity linking is to map the mention $m \in M$ to the $e \in E$ in the knowledge base $G$. }

\textbf{Few-Shot Entity Linking} \textit{The few-shot entity linking is a specific entity link task aimed at accurately linking mentions of free texts to an existing knowledge base with minimal labeled data. Formally, for a knowledge base $G$ which has $N$ entities, we only have $M$ linking examples  $\{(m_i,e_i),1\leq i \leq M\}$, where $M\ll N$. }

In the above definition of \textbf{Entity Linking}, the mentions that emerged in the texts are predefined by various methods, including entity recognition and other synthetic approaches. The knowledge base here refers to the collection of the entities, their semantic relations, and attributes. Generally, entity linking is the second step after the entity recognition process and can facilitate many tasks. Actually, some mentions may not have corresponding entities in the knowledge base. In our study, we assume that all mentions have their corresponding entities in the knowledge base. Previous studies have shown that entity linking is challenging due to the variations of entity names and entity ambiguity. Many methods have been developed to solve these problems. However, existing entity linking methods require a huge amount of high-quality labeled data, which may be labor-intensive. Unfortunately, high-quality data is generally difficult to access, especially in some specific domains. For example, mentions in the medical domain need to be confirmed by medical experts. In such situation, merely little data can be available. Therefore, there is an urgent need to solve the entity linking problem in the few shot setting. In our paper, the few-shot entity task is aimed at linking the mentions of texts in specific domains without enough labeled data.

\subsection{Weak Supervision}
With sufficient high-quality data, supervised methods can gain comparable success on multiple tasks. In general, the input instance of the supervised learning methods contains two parts: one is the feature vector, and the other is the label. Due to the complexity of real-world learning tasks, high-quality labels are sometimes unavailable. In contrast, noisy and incomplete labels with limited information can be easy to access, which leads to the research of weakly supervised learning~\cite{zhou2018brief}. 

Generally, for supervised learning methods, the best model is trained on high-quality data, which is not realistic when  methods are deployed on essential conditions without enough labeled data. Formally, given input data $X=\{(x_0,y_0),\cdots,(x_m,y_m),x_{m+1},\cdots,x_{m+n}\}$, where $m$ is the number of labeled data and $n$ is the number of unlabeled data, the paradigm of weakly supervised learning first generate a 
 pseudo label $y^{'}_i$ for each unlabeled instance $x_i, m<i \leq m+n$ and reconstruct the training dataset as $X^{'} =\{(x_0,y_0),\cdots,(x_m,y_m),(x_{m+1},y^{'}_{m+1}),\cdots,(x_{m+n},y^{'}_{m+n})\}$. Based on the pseudo label, many supervised machine learning algorithms can be applied on the synthetic dataset $X^{'}$ to learn the predictive models. 

\subsection{Meta-Learning}
The meta-learning mechanism is an important architectural component of our few-shot entity linking system. In the following, we will present an uncomplicated overview and analysis of the mainstream approaches of meta-learning. Meta-learning is the scientific approach of systematically observing how different machine learning approaches perform on a wide range of learning tasks and then by learning from this experience, or metadata, to learn new tasks much faster than otherwise possible. Not only does this dramatically speed up and improve the design of machine learning pipelines or neural architectures, but it also allows us to replace hand-engineered algorithms with novel approaches learned in a data-driven way. 

Although the continuously evolving deep neural networks have achieved exciting results on various tasks in recent years, existing methods still suffer from diverse deficiencies. For example, humans can learn quickly from a tiny number of samples, while training a neural network with desired performance requires much more data, which leads to its inability to learn quickly. Meta-learning is the method used to solve this problem \cite{Hospedales2021meta}. \lee{In this paper, we mainly take advantage of the meta-learning methodology to adjust the weight of the synthetic data with a small number of samples in the target domain so that the model can quickly adapt to a new domain.} 

%% file: sec4-framework.tex
\section{Framework}\label{sec-framework}

\begin{figure*}
  \centering
  \includegraphics[width=0.95\linewidth]{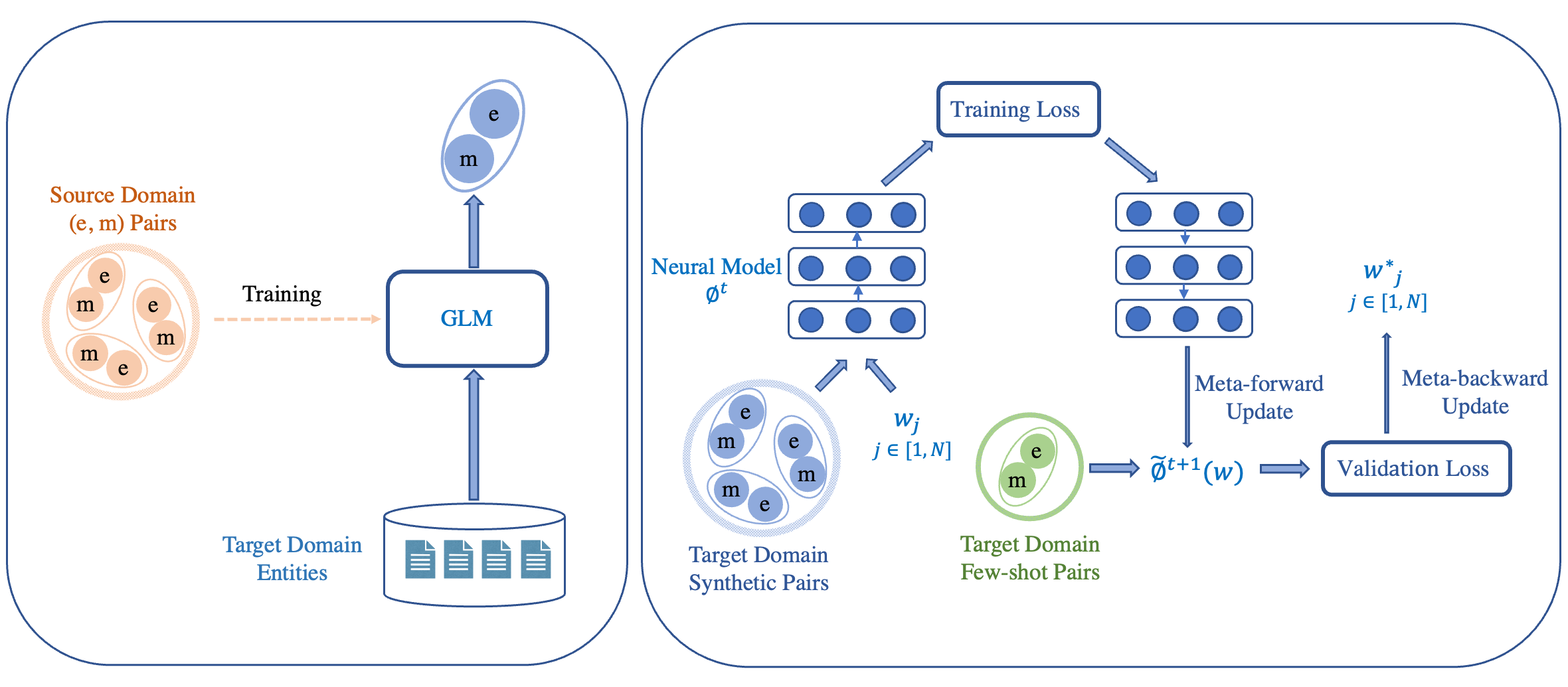}
  \caption{Illustration of our framework. The left part (a.k.a., Target Domain Mention Generation) focuses on generating large amount of synthetic supervision in target domain. The right part (a.k.a., \lee{Meta-Learning enhanced Entity Linking}) adopts a meta-learning method to reweight the synthetic data by leveraging the few-shot data in target domain, and learn a neural model based on weighted samples to solve the problem of entity linking.}
  \label{framework}
\end{figure*}

In this section, we introduce the whole framework of our proposed model, which is denoted as MetaBLINK. As shown in Figure~\ref{framework}, MetaBLINK contains two parts: ``Target Domain Mention Generation'' and ``\lee{Meta-Learning enhanced Entity Linking}''.

\subsection{Target Domain Mention Generation.~} We leverage the pre-training model T5~\cite{DBLP:journals/jmlr/RaffelSRLNMZLL20} to generate many mentions for the target domain in this part. In particular, considering the availability of labeled data in many source domains, we collect the entity-mention pairs(denoted as (e,m)) from these source domains and then use them for training the T5 model. In this way, each mention is generated from an entity. \lee{To further improve the performance of our model, we also inject knowledge of the target domain into T5 by an unsupervised denoising task.} Next, we can leverage the trained T5 model to generate mentions for the given entities in the target domain. Therefore, we can get sufficient synthetic entity-mention pairs for the target domain.

\subsection{\lee{Meta-Learning enhanced Entity Linking}:~} Generally, we can learn the entity linking model for the target domain based on the synthetic entity-mention pairs because these data can be regarded as labeled data. However, synthetic data includes both normal cases (i.g., true samples for the target domain) and bad cases (i.g., wrong/meaningless samples for the target domain). Therefore, our goal is to remove bad cases for a better entity linking model. To achieve so, we leverage the meta-learning approach to score each synthetic sample and then iteratively use the weighted samples to retrain the entity linking model. In particular, the meta-learning process needs the few-shot pairs in the target domain to update the weights of synthetic samples. To this end, we can get both better weighted samples and a better trained entity linking model.

%% file: sec4-methodology.tex
\section{Methodology}\label{sec-methodology}
In this section, we give detailed illustration about the MetaBLINK model. As mentioned before, we are aiming at  resolving the following questions:
i) how to generate ``synthetic  samples"  ; ii) how to use the meta learning to reweight synthetic samples for effectively training an entity linking model.

In particular, the three questions can be answered by the process of ``Target Domain Mention Generation'' , the process of ``\lee{Meta-Learning enhanced Entity Linking}'', respectively. 
\subsection{Target Domain Mention Generation}\label{Synthetic-Supervision}
To tackle labeled data lacking problem in the target domain, we first use heuristic methods to get some samples for weak supervision. After that, we adopt a natural language generation model (NLG) to rewrite the mentions to generate more effective samples, which we call ``synthetic samples" in the target domain. In this section, we demonstrate the whole procedure in detail.

The prime procedure of synthetic sample generation contains two parts: exact matching and mention rewriting. To reduce the bias introduced in generating labels on the few-shot domain, we first utilize the enormous unlabeled data and employ the exact matching methods to generate the data in the target domain. To further improve the quality of generated data, we adopt the T5 model to generate more semantic-like data samples.

\textbf{Exact Matching.}\ 
Following Le et al. \cite{le2019distant}, we generate training samples in the target domain using ``Name Matching". More specifically, if a mention's name matches an entity's title, we will link the mention to the corresponding entity in the exact matching step. Through this heuristic method, we can match massive samples in the target domain. Nevertheless, there is a fatal flaw in this way. Suppose we directly train a neural model on these samples. In such case, the powerful models will find a shortcut——straightforward measuring the similarity between mentions and entities' title to resolve the entity linking problem, rather than measuring the semantic similarity between the context and the entities' description. These trivial synthetic entity-mention pairs will result in the model being far from adequately trained, which will seriously affect the model's performance in the prediction process. \lee{Though the ``exact match" method ensures that mention text matches title text, we find that there are still some errors because an entity may have different mentions.}

\textbf{Mention Rewriting.}\
To alleviate the bias mentioned above, we leverage the T5 model. A pre-trained language generation model that can be applied to rewrite the mentions. In this section, we will describe how to train and inference through the T5 model. 

Initially, we will give a brief introduction of T5, and more details can be found in \cite{DBLP:journals/jmlr/RaffelSRLNMZLL20}. T5 is an encoder-decoder architecture based on the transformer, pre-trained on a large corpus named ``C4". It is a unified model for NLU and NLG tasks. By casting all  NLU and NLG tasks to a text-to-text format with different task prefixes, it can give corresponding outputs with the same model.

\begin{figure*}
  \centering
  \includegraphics[width=0.8\linewidth]{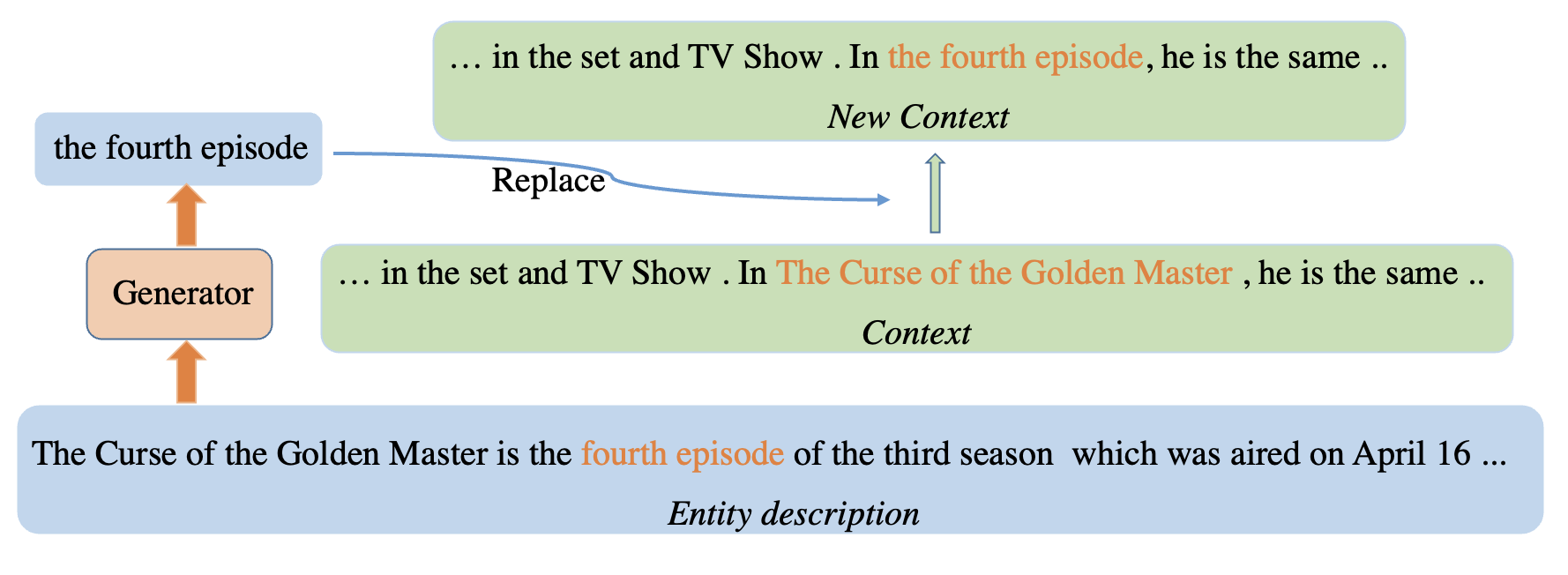}
  \caption{Example of mention rewriting. In the beginning, the mention ``The Curse of the Golden Master" is the same as the entity title. Then we feed the entity's description into the generator and it generates the new mention ``the fourth episode", which will replace the original mention. The new context and original entity description will form a new training pair.}
  \label{fig:generator}
\end{figure*}

In our model, we suppose the mention contains part of semantic information of the corresponding entities, so we add the prefix ``summarize:" to the entity's description to force the model to summarize the entity in a few words. \lee{We regard the mention rewriting procedure as a summarization task. The input is an entity’s description with "summarize:" prefix. The output is a mention. For example, the input is ``summarize: The Curse of the Golden Master is the fourth episode of the third season, which was aired on April 16 ..." The output of T5 is ``the fourth episode."} Figure~\ref{fig:generator} is an illustration of the rewriting procedure.

At first, we train T5 in the general domain:
\begin{equation}
m = T5 \ (\tau(e)) 
\end{equation}
where $e$ is the text description of an entity and $m$ is its corresponding mention. $\tau(e)$ is a function that adds the prefix ``summarize:" to an entity' s description.

Then we use the trained T5 to rewrite mentions in the target domain:

\begin{equation}
m' = T5 \ (\tau(e)) 
\label{T5:train}
\end{equation}

where $m'$ is the new mention, which will replace the original mention to form a new context. \lee{Note that the original mention has an impact on the generated mention. In the exact match step, mention and entity title are the same. In equation~\ref{T5:train}, entity title and entity description are used as input.}

\lee{We designed a fine-tuning strategy that allows T5 to learn semantic information of the target domain. Specifically, we masked the document text with specific sentinel tokens and forced T5 to recover the original text. Through such a procedure, T5 can adapt to a new domain without manual annotation. For example, the sentence ``Square Foot is a minifigure released in October 2015" can be masked as ``Square $<extra\_id\_0>$ is a $<extra\_id\_1>$ released in $<extra\_id\_2>$."}

\subsection{\lee{Meta-Learning enhanced Entity Linking}}
In this section, we first introduce the general process of leverage an deep learning model to solve the entity linking problem, where the model takes as input each synthetic sample (Sec.~\ref{entity-link-model}). Next, we leverage the meta learning mechanism to reweight training samples with the help of few real samples in target domain, which is also knows as the denosing process (Sec.~\ref{denosing}).

\subsubsection{Entity Linking Model} \label{entity-link-model}

 In the  weakly supervised dataset $\{(m_i,e_i)\}_{i=1}^N$ , each training sample  contains two elements, one is the mentions and their contexts, and the other is the entities  and their corresponding descriptions. Generally, for mention representation learning, the mention encoder $\mathop{ENCODER}^{m}$ takes the mention  and its context as input and output the dense representation of the mention. Similarly, for entity embedding, the entity encoder $\mathop{ENCODER}^{e}$ takes the entity and the description of the entity as the input and output the dense representation of the entity.

Formally, the encoding process of a mention $m_i$ and an entity $e_i$ can be computed in the following way: 

\begin{equation}
  \mathbf{m_i} =  ENCODER^{m}(m_i,\mathop{context}(m_i))
\end{equation}

\begin{equation}
  \mathbf{e_i} =  ENCODER^{e}(e_i,\mathop{desp}(e_i))
\end{equation}

After that, the match score of a mention  $m_i$ and an entity $e_i$ can be computed by the cosine distance of their representations:
\begin{equation}
  S(m_i,e_i) =   \mathbf{m_i} \cdot  \mathbf{e_i}
\end{equation}

To optimise the entity linking model,  the  loss function is designed to maximize the correct entity-mention pairs with respect
to the (randomly sampled) entities of the same
batch $E(B)$\cite{wu-etal-2020-scalable}. Formally, the loss function is defined as follows:

\begin{equation}
  l_i(m_i,e_i) =  -S(m_i,e_i) + log\sum_{e\in E(B) \backslash \{e_i\}}
  exp (S(m_i,e))
\end{equation}

\subsubsection{Denoising with Meta-Learning}\label{denosing}

Through exact matching and mention rewriting, we have a large number of training instances in the few-shot domain. And in the section \ref{entity-link-model}, we give a deep learning based entity linking framework, which can be used to train on the weakly supervised dataset. However, part of the instances are noisy and may deteriorate the entity linking model's performance in the target domain. Therefore, we need to find an effective way to select suitable instances for the training process. Previous studies have shown that meta-learning can be used to reweight the instances during the training process automatically and gained relative success in neural ranking~\cite{DBLP:conf/icml/RenZYU18}. Therefore, we adopt a meta-learning method to reweight these synthetic data in this section. 

Formally, to find the best weights for each sample, we need a few target domain samples(seed dataset) to initialize the training process of the weight. Therefore, Given the  target domain samples $\{(m_i,e_i),1\leq i \leq M\}$ and the synthetic target domain samples $\{(m'_j,e'_j),1\leq j \leq N \}(M\ll N)$, the optimization ``objective" of entity linking with weighted samples are as the follows:

\begin{equation}
\begin{split}
     \phi(w) = \mathop{\arg\min}\limits_{\phi}\sum_{j=1}^N w^*_jl_j(m'_j,e'_j;\phi)
     \\
     s.t. \quad w^* =  \mathop{\arg\min}\limits_{w}\sum_{i=1}^M l_i(m_i,e_i;\phi(w))
\end{split}
\label{equa:meta-objective}
\end{equation}
where $l(m,e;\phi)$ is the loss on $(m,e)$.

\lee{There are two parameters to be optimized in the training process: the model parameter $\phi$ and the training instance weights $w^*$.} To solve the equation~\ref{equa:meta-objective}, following the previous work~\cite{DBLP:conf/icml/RenZYU18},\cite{sun2020meta}, we divided the optimization stage into three iterative process: ``Meta-forward Update", ``Meta-backward Update" and ``Training with New Weight".

\textbf{Meta-forward Update.}
Suppose the initial weight for synthetic samples is $w$. The meta-forward update optimizes a pseudo neural model parameter $\hat{\phi}$:
\begin{equation}
    \hat{\phi}(w) = \mathop{\arg\min}\limits_{\phi}\sum_{j=1}^N w_jl_j(m'_j,e'_j;\phi)
\end{equation}

Note that finding the optimal $\hat{\phi}$ can be a time-consuming procedure, so we can perform one gradient descent optimization step with a mini-batch sample. Supposing $\phi^{t}$ is the parameter at step $t$ and there exist $n$ samples in a mini-batch, then $\hat{\phi}^{t+1}$ can be calculated as:
\begin{equation}
    \hat{\phi}^{t+1}(w) = \phi^{t}-\alpha\frac{\partial}{\partial(\phi^t)}\sum_{j=1}^n w_jl_j(m'_j,e'_j;\phi^t)
\end{equation}
where $\alpha$ is the learning rate.

\textbf{Meta-backward Update.}
In meta-backward update, we can use $\hat{\phi}(w)$ above to find the optimal weight $w$:
\begin{equation}
    w^* = \mathop{\arg\min}\limits_{w}\sum_{i=1}^M l_i(m_i,e_i;\hat{\phi}(w))
\end{equation}

Besides, here we can use one gradient descent step to optimize $w$:

\begin{equation}
    w^*_{j} = w_j - \alpha\frac{\partial}{\partial(w_j)}\sum_{i=1}^m \frac{1}{m}l_i(m_i,e_i;\hat{\phi}^{t+1}(w))
\end{equation}

Following Ren et al. \cite{DBLP:conf/icml/RenZYU18}, in practice we set the initial weight $w$ as zero, so the formula can be changed as:
\begin{equation}
     \widetilde{w}_{j} = -\eta\frac{\partial}{\partial(w_j)}\sum_{i=1}^m \frac{1}{m}l_i(m_i,e_i;\hat{\phi}^{t+1}(w)) \left. \right|_{w_j=0}
\end{equation}

Furthermore, to prevent negative weight and make training stable, we drop the negative weight and add normalization step:
\begin{equation}
     \hat{w}_{j} = max(0,\widetilde{w}_j)
\end{equation}

\begin{equation}
    w^*_j =\frac{\hat{w}_{j}}{(\sum_{p=1}^n\hat{w_p})+\delta(\sum_{p=1}^n\hat{w_p})} 
\end{equation}
where $\delta(\sum_{p=1}^n\hat{w_p})$ equals 1 if and only if $\sum_{p=1}^n\hat{w_p} $ equals 0.

\textbf{Training with New Weight.}
After getting the new weight $w^*$, the neural model can be optimized using standard back propagation:
\begin{equation}
    \phi^{t+1} = \mathop{\arg\min}\limits_{\phi^t}\sum_{j=1}^n w^*_jl_j(m'_j,e'_j;\phi^t)
\end{equation}

In summary, we first calculate the weight of synthetic data based on their impact on few-shot data in the target domain, and then we use the weighted loss to update the neural model. 

We depict the training procedure step-by-step in pseudo-code in Algorithm \ref{alg1}, where $D_f$ denotes the synthetic data and $D_g$ denotes the few-shot data. And the full training process of MetaBLINK is proposed in Algorithm \ref{alg2}.

\begin{algorithm}[ht!]
\setstretch{1.15}
  \caption{Learning to Reweight Synthetic data}
  \label{alg1}
  \normalsize
  \begin{algorithmic}[1]
     \Require Synthetic dataset $D_f$ and seed dataset $D_g$, training batch size   $n$, meta batch size $m$, initial model parameter $\phi_0$
     \Ensure model parameter at time step t $\phi_T$
     \For{t = 0...T-1} 
     \Statex \quad // sample a synthetic batch of size $n$ from synthetic 
     \Statex \quad dataset 
     \State  $\{m_f,e_f\} \leftarrow SampleBatch(D_f,n)$
     \Statex  \quad // sample a seed batch of size $m$ from the seed set 
     \State  $\{m_g,e_g\} \leftarrow SampleBatch(D_g,m)$
     \Statex  \quad // init the weight $w$  with zero weight and compute the 
     \Statex  \quad loss on the synthetic batch
     \State  $w \leftarrow 0; l_f \leftarrow \sum_{j=1}^{n}w_{j}L(m_{f,j},e_{f,j})$
     \Statex  \quad // compute the gradient of $\phi_t$ with respect to the loss
     \Statex  \quad  of the synthetic batch
     \State  $\nabla \phi_t \leftarrow Backward(l_f,\phi_t)$
     \State  $\hat{\phi}_t \leftarrow \phi_t - \alpha\nabla\phi_t$
     \Statex  \quad // use the updated model to compute loss on seed batch:
     \State $l_g \leftarrow \frac{1}{m}\sum_{i=1}^{m}L(m_{g,i},e_{g,i};\hat{\phi}_t)$
     \Statex  \quad // calculate the derivative of the loss with respect to the
     \Statex  \quad  initial weight:
     \State  $\nabla w \leftarrow Backward(l_g,w)$
     \Statex \quad // determine the weight of synthetic batch:
     \State  $\widetilde{w}\leftarrow max(-\nabla w,0);w\leftarrow \frac{\widetilde{w}}{\sum_{j}\widetilde{w}+\delta(\sum_{j}\widetilde{w})}$
     \Statex  \quad // truely optimize the model with weighted loss:
     \State  $\hat{l}_f\leftarrow\sum_{j=1}^{n}w_jL(m_{f,j},e_{f,j};\phi_t)$
     \State  $\nabla \phi_t \leftarrow Backward(\hat{l}_f,\phi_t)$
     \State  $\phi_{t+1}\leftarrow Optimize(\phi_t,\nabla \phi_t)$
    \EndFor
    \end{algorithmic}
    \end{algorithm}

\begin{algorithm}
\setstretch{1.25}
  \caption{MetaBLINK Training Framework}
  \label{alg2}
  \normalsize
  \begin{algorithmic}[1]
    \Require Source Domain dataset $D_s$, seed dataset $D_g$, T5, initial MetaBLINK model $\phi_0$
    \Ensure best MetaBLINK model $\phi$
    \State Generating target domain data $D_f$, using Exact Matching algorithm using target   domain entities.
    \State Training T5 with summary task and and augmenting the target domain dataset $D_f$ by rewriting mentions.
    \State Training MetaBLINK using Algorithm \ref{alg1} with $D_f$, $D_g$ and $\phi_0$.
  \end{algorithmic}
\end{algorithm}

%% file: sec5-implementation.tex
\section{Implementation}\label{sec-implementation}
In this section, we present the implementation of MetaBLINK in detail. 

For the reason that MetaBLINK is based on the BLINK\cite{wu-etal-2020-scalable}, most hyper-parameters are set as the same as BLINK: bi-encoder and cross-encoder are optimized with Adam, with the learning rate of 2e-5. The batch size is 128 for the bi-encoder. For cross-encoder, due to meta-learning progress consumes twice as much as initial memory, we can only set the batch size to one. The best model is selected based on the performance of the development set. For the synthetic data generator T5, we select the training domains of the zero-shot entity linking dataset as our training source and train it for 20 epochs with a learning rate of 2e-5. Our model is trained on 2 NVIDIA GeForce RTX 3090 GPUs.

In the few-shot scenario, we directly select 50 samples from the dataset as the seed for meta-learning. We obtain the seed samples in the zero-shot scenario by filtering the synthetic data and using a heuristic self-match method. More details are presented in the next section. 

%% file: sec6-evaluation.tex

\vspace{-.5em}
\section{PERFORMANCE EVALUATION}\label{experiments}\label{sec:experiment}

\begin{table*}[h]
    \centering
        \caption{Examples exhibiting errors made by the model trained on ``Exact Match" data. The red words are mentions and the orange words in the descriptions are keywords that also occur in the mention's context. The entities colored by gray are wrong answers.}
\resizebox{1.\textwidth}{!}{
    \begin{tabular}{m{0.5\columnwidth}m{0.5\columnwidth}m{0.5\columnwidth}<{\centering}m{0.5\columnwidth}<{\centering}}
    \toprule
        \textbf{Mention} & \textbf{Corresponding Entity}  & \textbf{Exact Match} &\textbf{ Syn}\\
    \midrule
    
... misty suggested that she partnered with the player and that jack partnered with \textcolor{red}{taku}. In the final event, misty asked the player to ... & \textbf{Yosuke Adachi} ... his name is changed to \textcolor{orange}{Taku} ... Taku and Jack Atlas duel against the \textcolor{orange}{Player} and \textcolor{orange}{Misty} ...

&\textcolor{gray}{Taku} \ & \textbf{Yosuke Adachi}   \\ \midrule

 ... sartorius wins easily without even using one turn and he obtains the controls to misgarth' s \textcolor{red}{satellite}. jaden wonders what kind of duels are in store for the rest of the ...  & \textbf{SORA (satellite)} SORA is the \textcolor{orange}{satellite} belonging to the country of Misgarth. \textcolor{orange}{Sartorius} was awarded the keys to this satellite by Prince Ojin upon winning their \textcolor{orange}{Duel}...
& \textcolor{gray}{Satellite (series)} \ & \textbf{SORA (satellite)}   \\ \midrule

... when \textcolor{red}{his father} came to visit, he gave tomoya a zombire figure, telling him that this is zombire and he is the strongest hero in America ... & \textbf{Mr. Hanasaki} Mr. Hanasaki, is Tomoya Hanasaki's \textcolor{orange}{father}. Mr. Hanasaki is a \textcolor{orange}{``Zombire"} fan. He works in \textcolor{orange}{America} ...
& \textcolor{gray}{The Father We Remember}  & \textbf{Mr. Hanasaki} \\ \midrule

    \end{tabular}
    }
    \label{tab:my_label}
\end{table*}

In this section, we conduct comprehensive experiments on the few-shot entity linking task to evaluate the effectiveness of our model. To start with, we introduce the related datasets and illustrate implementation details. Afterward, we conduct a series of analyses of the experimental results. 

\subsection{Experimental Setup}\label{sec:experiment:setup}
\textbf{Dataset} We conduct extensive experiments on the benchmark used in \cite{DBLP:conf/acl/LogeswaranCLTDL19}  to evaluate the zero-shot entity linking task. \lee{The benchmark dataset is collected from fandom\footnote{https://www.fandom.com/explore-zh}, and the detailed statistics are shown in Table \ref{Zero-Shot Entity Linking Dataset}.} 
Based on the string overlap between mention and its corresponding entity title, the samples can be divided into four categories: \textit{High Overlap}, \textit{Multiple Categories}, \textit{Ambiguous Substring}, and \textit{Low Overlap}. \textit{High Overlap} refers to the set of samples whose mention text is the same as the title text; \textit{Multiple Categories} is the set of samples whose title text is the mention text followed by a disambiguation phrase, \textit{Ambiguous Substring} refers to the set of samples whose mention text is a substring of the title text, while \textit{Low Overlap} is the set of samples not belonging to the above three categories.

\begin{table}[ht]
  \centering
  \caption{Zero-Shot entity linking dataset.}
  \begin{tabular}{lll}
  \toprule
  Dataset &  Domain  &   Entities\\ \hline
  \multirow{8}{*}{\#Train}     
  & American Football  &  31,929\\ 
  & Doctor Who & 40,821 \\ 
  & Fallout & 16,992 \\ 
  & Final Fantasy & 14,044 \\ 
  & Military &104,520\\
  & Pro Wrestling &10,133\\
  & StarWars&87,056\\
  & World of Warcraft &27,677\\ \midrule
  \multirow{4}{*}{\#Dev}
  &Coronation Street & 17,809\\
  & Muppets &21,344\\
  & Ice Hockey & 28,684\\
  & Elder Scrolls&21,712\\ \midrule
  \multirow{4}{*}{\#Test} 
  & Forgotten Realms & 15,603\\
  & Lego & 10,076\\
  & Star Trek & 34,430\\
  & YuGiOh & 10,031\\
\bottomrule
  \end{tabular}
  \label{Zero-Shot Entity Linking Dataset}
\end{table}

To evaluate the performance of  MetaBLINK under the few-shot circumstances, we further split the samples in each of the four test domains into training/development/test datasets, shown in Table \ref{few-shot dataset}. \lee{We selected the number of samples among  \{10, 20, 30, ..., 100\}.} Specifically, we split off 50 samples as the training set, 50 samples for the development set and keep the rest as the test set.

\begin{table}[ht]
  \centering
  \caption{Few-shot entity linking dataset.}\scalebox{1}{
  \begin{tabular}{ccccc} \toprule
  Domain &  \ \#Train & \ \#Dev  & \ \#Test & \\ \midrule
  Forgotten Realms     & 50  & 50  & 1,100 \\ 
  Lego    & 50 & 50 & 1,099  \\ 
  Star Trek    & 50 & 50 & 4,127  \\ 
  YuGiOh    & 50 & 50 & 3,274  \\ \bottomrule
  \end{tabular}}
  \label{few-shot dataset}
\end{table}

\textbf{Evaluation protocol} 
In the experiment phase, entity linking can be divided into two stages: candidate generation and candidate ranking. \lee{We use recall and normalized accuracy (N.Acc.) to measure the performance of the two stages separately. Note that normalized accuracy only considers samples for which the right entity is within the candidates retrieved by the first stage. We use unnormalized accuracy (U.Acc.) to measure the performance of the whole procedure, which is the product of recall and normalized accuracy.}

\subsection{Few-Shot Entity Linking in Specific Domain}
Firstly, we design comprehensive experiments to demonstrate the effectiveness of MetaBLINK when only minimal samples can be available in specific domains. We denote the small amount of data in the target domain as seed. \lee{For the synthetic data, we denote the data generated by T5 that is trained only on supervised data on the source domains as syn, and the data generated by T5 that is trained both on supervised data on the source domains and unsupervised data on the target domain as syn*.} \lee{Note that the seed was obtained in different ways under different settings. Under the few-shot setting, we directly split off a few samples as the seed. Since there is no available data on the target domain under the zero-shot domain transfer setting, we use heuristic methods to generate the seed. All syn refers to the same data in the paper. The number of generated data is related to the data scale of a certain domain. Specifically, the numbers of generated data on the Forgotten Realms/Lego/Star Trek/YuGiOh domains are 1656, 1986, 1939, and 1103 respectively.} The baselines we compare with MetaBLINK are:

(1) Name Matching, a naive method proposed by Riedel et al.~\cite{riedel2010modeling} for relation extraction: if a mention matches an entity's name, then it is regarded as linked to that entity.

(2) BLINK, a strong baseline for zero-shot entity linking task, proposed by Wu et al.~\cite{wu-etal-2020-scalable}, which adopts the pre-trained model for both candidates generating stage and candidate ranking stage. To apply BLINK in the few-shot environment, we have made some modifications by training BLINK on the small amount of data in the target domain (seed) and adopt it as our baseline. We also train BLINK on different data sources (syn and syn+seed) for exhaustive comparison.

(3) DL4EL, a denoising method proposed by Le et al.~\cite{le2019distant}, assumes a noise ratio and uses KL divergence to force the model to select high-quality data. We refer readers to Le et al.~\cite{le2019distant} for detailed description. Note that DL4EL's denoising procedure is in-batch, while the cross-encoder's batch size is very small, so we only apply it to the bi-encoder. \lee{DL4EL can be applied in almost any architecture, making the model automatically learn to distinguish high-quality data from low-quality data and give larger weights to high-quality data. Morespecifly, we adopted the same bi-encoder and cross-encoder architecture as BLINK in implementing DL4EL.}

\begin{table*}[ht]
  \centering
  \caption{U.Acc. on Forgotten Realms domain and Lego domain. Syn represents the synthetic data and Seed represents the few-shot data in target domain.}\scalebox{1}{
  \begin{tabular}{ll|rrl|rrl}
  \toprule
  & & \multicolumn{3}{c}{Forgotten Realms} &\multicolumn{3}{c}{Lego}\\\hline
  Method & Data & R@64 & N.Acc. & U.Acc.& R@64 & N.Acc. & U.Acc. \\ \midrule
Name Matching \cite{riedel2010modeling} & - & - & - & 19.64 & - & - & 12.37\\
BLINK \cite{wu-etal-2020-scalable} & Seed   & 35.27 & \textbf{59.02} & 20.82      &  52.68 & 45.60 & 24.02   \\
\midrule
BLINK & Syn &  63.82 & 40.33 & 25.74 &  72.88 & 28.59 & 20.83   \\
BLINK & Syn+Seed   &  66.75 & 54.10 & 36.11  &  75.52 & 48.80 & 36.85  \\
DL4EL \cite{le2019distant} & Syn+Seed & 66.71 &  54.10 & 36.09 & 75.12 &  48.80 & 36.65  \\
MetaBLINK & Syn+Seed & 68.91 & 56.33 & 38.82 & 77.80 & 50.18 & 39.04\\
MetaBLINK & Syn*+Seed & \textbf{69.56} & 56.27 & \textbf{39.14} ($\uparrow18.32$) & \textbf{78.17} & \textbf{50.65} & \textbf{39.59} ($\uparrow15.57$) \\
  \bottomrule
  \end{tabular}
  }
  \label{few-shot res1}
\end{table*}

\begin{table*}[ht]
  \centering
  \caption{U.Acc. on Star Trek domain and Yugioh domain.}\scalebox{1}{
  \begin{tabular}{ll|rrl|rrl}
  \toprule
  & & \multicolumn{3}{c}{Star Trek} &\multicolumn{3}{c}{Yugioh}\\\hline
  Method&Data & R@64 & N.Acc. & U.Acc.& R@64 & N.Acc. & U.Acc. \\ \midrule
Name Matching \cite{riedel2010modeling}&-&-&-& 12.12 & - & - & 7.88\\
BLINK \cite{wu-etal-2020-scalable}&Seed   &21.57&\textbf{37.08}& 8.00      &  35.00 & 37.70 & 13.20   \\
\midrule
BLINK&Syn &  55.61 & 21.31 & 11.85   &  55.77 & 22.84 & 12.74  \\
BLINK&Syn+Seed   &  59.82 & 32.14 & 19.23     &  60.72 & 35.11 & 21.32 \\
DL4EL \cite{le2019distant} &Syn+Seed & 59.91 &  32.14 & 19.26  & 59.20 &  35.11 & 20.79  \\
MetaBLINK &Syn+Seed  &61.09&34.51& 21.08  &  
60.72 &  37.58 & 22.82\\
MetaBLINK &Syn*+Seed  &\textbf{61.41}&34.63& \textbf{21.27} ($\uparrow9.15$) &  \textbf{61.28} &  \textbf{38.02} & \textbf{23.30} ($\uparrow10.10$) \\
  \bottomrule
  \end{tabular}
  }
  \label{few-shot res2}
\end{table*}

Results are shown in Table~\ref{few-shot res1} and Table~\ref{few-shot res2}. According to the results, we have the following observations:

(1) Since the majority of samples belong to \textit{Low Overlap} type, the naive name matching method which only relies on shallow surface text doesn't perform well, indicating that we should integrate context and description information into the model.

(2) Despite BLINK's powerful reading comprehension ability, merely using the seed set cannot support it to resolve existing problems caused by insufficient labeled data, which reveals the contradiction between the huge-scale model parameters and the minimal number of samples. Comparing it to BLINK trained on the synthetic data, we find that high-quality seed data lead to better cross-encoder accuracy, while synthetic data enhances the bi-encoder significantly. It indicates that cross-encoder is more ``quality sensitive," while a large amount of data can help bi-encoder learn useful patterns, which is important for recall.

(3) DL4EL doesn't bring improvement as expected. In contrast, it leads to a decline in algorithm performance sometimes. \lee{As there is no obvious ``bad data" in synthetic data (such as spelling mistakes), DL4EL cannot recognize ``bad data" based on simple data features.} At the same time, due to the lacking of high-quality seed data, the model cannot distinguish useful data well. In contrast, MetaBLINK can combine the strength of synthetic data and seed data. Moreover, using the seed data as extra supervision, the model can make full use of the synthetic data and achieve the best performance, which gains significant improvements compared to name matching or directly fine-tuned models.

(4) \lee{Experiments showed that this procedure could slightly improve T5's rewriting ability. MetaBLINK gains further improvement by leveraging using syn* data compared to syn data. The experimental results indicate that the additional unsupervised denoising training step can improve the rewriting ability of T5 via generating more fluent mentions with fewer errors. }

\subsection{Zero-Shot Domain Transfer}
In this section we evaluate our method on a more challenging setting: zero-shot entity linking. Logeswaran et al. \cite{DBLP:conf/acl/LogeswaranCLTDL19} present the zero-shot entity linking task, where they train the model on 8 training domains and then directly evaluate the model on 4 testing domains without the aid of any in-domain samples. \lee{We didn't put syn + seed because the 8 training domains already have high quality and plenty of data. Note that introducing the term general domain is to distinguish train set from specialized target domains. In addition, we found that BLINK pretrained on Wikipedia underperformed BLINK trained only on the train set.}

First, we use heuristic methods to generate the seed set instead of directly splitting it from the original dataset.
Specifically, we generate high-quality seed data in two strategies:

(1) Filtering the synthetic data by some rules, such as correct spelling, no overlapping between mention and entity title, etc.

(2) Self-match: \textit{Multiple  Categories} type occupies a large proportion in the dataset, while it is rare in our synthetic data. To fill this vacancy, we focus on entities whose title contains a disambiguation phrase. We find the mention in the entity's own description which is the same as the title (excluding the disambiguation phrase). The results are reported in Table~\ref{zero-shot res}.

\begin{table*}[h]
  \centering
  \caption{U.Acc. on four domains when doing zero-shot domain transfer.}\scalebox{1}{
  \begin{tabular}{ll|cccc}
  \toprule
   Method & Data & Forgotten Realms &Star Trek &Lego &Yugioh\\ \midrule
 BLINK &-&84.11 & 74.45  &  72.22 & 66.30 \\ \midrule
BLINK&Seed     &84.60 & 74.51 &  73.51 & 68.80 \\
MetaBLINK &Syn+Seed  & \textbf{84.81}  & \textbf{74.54} &  \textbf{74.11} &  \textbf{69.50}  \\
  \bottomrule
  \end{tabular} }
  \label{zero-shot res}
\end{table*}

The experimental results show that our model can achieve better performance on testing domains compared to BLINK. Moreover, the results are diverse across domains. In the Forgotten Realms domain and Star Trek domain, MetaBLINK only achieves little progress, while in the Lego domain and Yugioh domain, it leads to obvious improvement. We suppose it is because the ``gaps" between the general domain and testing domains are different. For example, the gap between the Star Trek domain and the general domain is small. Thus, it is tough to get improvement on Star Trek by leveraging pre-trained models.

\begin{table}[ht]
  \centering
  \caption{Gaps between general domain and four testing domains measured by U.Acc. difference.}\scalebox{0.92}{
  \begin{tabular}{l|cccc}
  \toprule
   Method&  Forgotten Realms &Star Trek &Lego &Yugioh\\ \midrule
   BLINK      & 85.19 & 74.49 & 71.79 & 65.61 \\
   BLINK+FT   &  88.55 & 77.04 & 78.46 &  73.08 \\
   GAP & 3.36 & 2.55 &6.67 &7.47 \\
  \bottomrule
  \end{tabular} }
  \label{gap}
\end{table}

To demonstrate our hypothesis, we design a tiny experiment to test the ``gap".  Specifically, for every testing domain, we select 500 samples for training, allowing models to be exposed to domain knowledge sufficiently. Then we regard the improvement of accuracy as the ``gap" between this domain and general domain.
Results are shown in Table\ref{gap}. BLINK+FT in the table means BLINK trained on the 500 samples.The results show that the gap between the general domain and the Forgotten Realms/Star Trek domain is relatively small, indicating that the model trained on the general domain already acquires the abilities required in these domains. On the other hand, the gap in Lego and Yugioh domains is relatively large, indicating that the model has much potential. Correspondingly, our MetaBLINK can achieve 2-4\% improvement in the Lego domain and Yugioh domain, demonstrating that our method can help models bridge the gap between different domains.

\lee{We summarize our design thought behind the experiment for zero domain transfer as follows: Table VII shows the results of MetaBlink in four target domains. The improvement on Forgotten Realms and Star Trek domains is not obvious. We suspect this may be because the gaps between these two domains and the general domain are relatively small. On this basis, we experimented with Table VIII to calculate the gap between target and general domains, and the experimental results confirmed our conjecture. Since Table IX is to verify MetaBLINK’s domain transferability, we abandoned Forgotten Realms and Star Trek domains for the zero-shot domain transfer experiment.}

In the aforementioned experimental phase, we utilize synthetic data as our training source, while in a zero-shot setting, we can take advantage of general domain data. Table\ref{different sources} shows the domain transfer results using different training sources.  

\begin{table}[h]
  \centering
  \caption{U.Acc. on two domains when doing zero-shot domain transfer using different training sources.}\scalebox{0.92}{
  \begin{tabular}{ll|lll}
  \toprule
   Method & Data &Lego &Yugioh&Avg\\ \midrule
 BLINK &- &   72.22 & 66.30 & 69.26 \\
BLINK&Seed     &   73.51 & 68.80 & 71.16\\
MetaBLINK &Syn+Seed  &  74.11 &  69.50 &71.81\\
MetaBLINK &General+Seed  &  74.82 &  68.90 &71.89\\
MetaBLINK &General+Syn+Seed &\textbf{74.90} &  69.52& 72.21
\\
MetaBLINK &General+Syn*+Seed &\textbf{74.90} &  \textbf{69.55} & \textbf{72.23}
\\
  \bottomrule
  \end{tabular} }
  \label{different sources}
\end{table}

From the results, we can find that the general domain data and synthetic data can both improve the transferability of model. We suppose the general domain data has higher quality while synthetic data introduces more domain-specific knowledge, \lee{and the distinction between syn and syn* data can be neglected when having plenty of general domain data.} Jointly using different sources can lead to nearly 3 points average improvement.

\subsection{The Effect of Meta Learning}
\begin{figure}[h]
  \centering
  \includegraphics[width=0.8\linewidth]{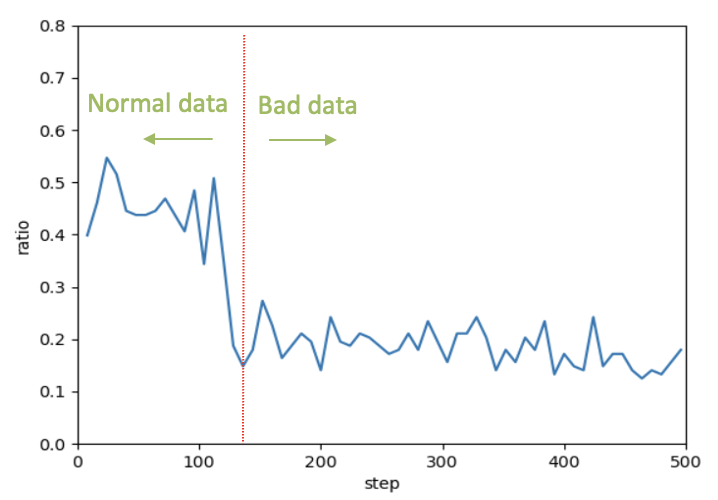}
  \caption{Effectiveness of meta learning: we firstly generate bad training data by linking mentions to random entities, then we record selecting ratio of different data source when supervised by few-shot data in target domain. The experiment is done on bi-encoder, Yugioh domain. }
  \label{meta learning res}
\end{figure}

In this section, we confirm the necessity of using meta-learning. We generate bad samples by linking mentions to random entities and test whether MetaBLINK can distinguish bad data from normal data. Figure \ref{meta learning res} shows the result, with the y-axis representing the selected ratio. We can find that by the supervision of the seed set, the model select nearly 50\% of the normal data, but only 20\% of the bad data, which indicating meta-learning plays an important role in the whole procedure.

\subsection{The Effect of Mention Rewriting}
In this part, we further design a series of additional experiments to investigate the effectiveness of mention rewriting on generating synthetic data. Concretely, we use Exact Match to represent the synthetic data that is generated only based on the strategy of exact matching. \lee{Firstly, we train BLINK on the different data sources and compare their corresponding linking accuracy.} The results are shown in Table \ref{effectiveness of synthetic data}. As expected, the syn data can effectively alleviate the bias and improve performance both on recall and ranking accuracy, \lee{and syn* achieves slight improvement in most cases by using an additional text denoising procedure.} Table \ref{tab:my_label} shows some typical errors made by the model trained on ``Exact Match" data.

\lee{Besides linking performance, we want to statistically analyze the difference between T5 generated data and Exact Match data. Specifically, we experiment to demonstrate that T5 generated data is closer to natural data distribution than Exact Match data. We introduce the ROUGE\footnote{https://github.com/huggingface/datasets/blob/master/metrics/rouge/rouge.py} metric to compare the golden mentions' similarity with T5 generated mentions and exact match mentions respectively. The golden mentions are sampled from the dataset of a specific domain. ROUGE metric is often used in the field of machine translation to evaluate the quality of abstract generation. As the length of generated mentions is relatively short, we adopted ROUGE-1 as the primary metric here. Table \ref{f1 score of different methods} shows the F1 score between golden mentions and different data sources. The results prove that the mentions generated by T5 are closer to the data distribution of the target domain. 
}
\begin{table*}
  \centering
  \caption{Effectiveness of mention rewriting.}
  \begin{tabular}{lllllllll}
    \toprule
    Domain    &\multicolumn{2}{c}{Lego} & \multicolumn{2}{c}{Yugioh}&\multicolumn{2}{c}{Forgotten Realms}&\multicolumn{2}{c}{Star Trek} \\\midrule
    Metric & R@64 & N.Acc. & R@64 & N.Acc.& R@64 & N.Acc. & R@64 & N.Acc.\\ \midrule
    Exact Match & 72.07 & 25.76 & 49.54 & 20.56 & 60.08 & 38.46& 54.22 & 20.74\\
    Syn  &72.88& 28.59  &55.77 & 22.84&63.82&\textbf{40.33}&55.61&21.31 \\
    Syn*  & \textbf{73.21} & \textbf{29.03}  & \textbf{56.32} & \textbf{23.36}&\textbf{64.61}&40.20&\textbf{55.71}&\textbf{21.36} \\
    \bottomrule
  \end{tabular}
  \label{effectiveness of synthetic data}
\end{table*}

\begin{table}
  \centering
  \caption{F1 score of ROUGE-1 using different methods.}\scalebox{0.92}{
  \begin{tabular}{lcccc}
    \toprule
    Domain    &\multicolumn{1}{c}{Lego} & \multicolumn{1}{c}{Yugioh}&\multicolumn{1}{c}{Forgotten Realms}&\multicolumn{1}{c}{Star Trek} \\\midrule
    Exact Match  & 33.70 & 38.01 & 40.18  &28.85\\
    Syn & 42.91 & 45.90 &42.26&33.98 \\
    Syn*  & \textbf{43.96} &  \textbf{46.56}  &\textbf{42.98}&\textbf{34.03} \\
    \bottomrule
  \end{tabular}}
  \label{f1 score of different methods}
\end{table}

%% file: sec2-background_and_overview.tex
\section{RELATED WORK}\label{related-work}

We address the problem of few-shot entity linking. Prior attempts to make up insufficient annotation data are all subject to one or more of the following limitations:

\begin{itemize}
    \item The process of introducing external knowledge will inevitably incur additional costs, and introducing external information may even introduce new errors.
    \item They depend excessively on external data, the effectiveness of the algorithms is directly affected by the quality of the external data, and the their utility is indispensable restricted.
    \item Neither can they generate the knowledge of the target domain nor improve the data quality by making adjustments according to the target domain.
\end{itemize}

Based on the techniques leveraged, existing few-shot entity linking optimization strategies can be divided into two categories: external information enhanced strategy and pre-trained model-based strategy. In this section, we briefly summarize the related works from these two aspects: external information enhanced entity linking and pre-trained model-based entity linking.  

\subsection{External information enhanced entity linking}
External information enhanced methods take advantage of different sources of external information to improve the quality of data, including manually curated mention tables~\cite{ganea-hofmann-2017-deep}, incoming Wikipedia link popularity~\cite{eshel-etal-2017-named}, and Wikipedia entity categories~\cite{gillick-etal-2019-learning}. Among external information sources, Wikipedia has gradually gained widespread attention from academia. Though external source information can provide supplementary semantic information, these methods inevitably have serious shortcomings. Firstly, introducing external information may introduce new errors; secondly, the performance of the algorithm is directly affected by the quality of the external data, and the applicability of the algorithm is indispensable restricted. 

There are recent works demonstrating that fine-grained entity typing information enhances the linking progress~\cite{raiman2018deeptype, onoe2020fine, khalife-vazirgiannis-2019-scalable}.~\cite{DBLP:conf/acl/LogeswaranCLTDL19} proposed the few-shot entity linking task. They use cross encoders for entity ranking, but rely on traditional IR-techniques for candidate generation and do not evaluate on large-scale benchmarks such as TACKBP~\cite{gillick-etal-2019-learning} show that dense embeddings work well for candidate generation. However, they do not involve pre-training or improving the quality of the generated data but include external category labels in their bi-encoder architectures, which limits their linking to entities in Wikipedia. Our approach can be seen as combining the backbones of existing works and demonstrate for the first time that pre-trained few-shot architectures are both highly accurate and computationally efficient at scale.

\subsection{Pre-trained model-based entity linking}
The study of pre-trained model-based few-shot entity linking is still in its infancy. While some pioneering works have been proposed, they used the BERT-based model directly to represent entities and related information. Wu et al. present a two-stage BERT-based linking algorithm~\cite{wu-etal-2020-scalable}. The first stage does retrieval in a dense space defined by a bi-encoder that independently embeds the context and the entity descriptions. Then each candidate is re-ranked with a cross-encoder that concatenates the mention and entity text. However, neither can it generate knowledge of the target domain nor make adjustments according to the target domain to resolve the ambiguity of entities. It is the most relevant work to us and keeps the state-of-the-art performance in this branch. 

%% file: sec7-future_work.tex
\section{Future Work}\label{sec-future_work}
In future work, we will work on investigating the remaining bottlenecks in MetaBLINK’s performance. We next plan to focus on the process of noise data generation; in addition, MetaBLINK might benefit from meta-learning mechanism improvements and supplementary information. In MetaBLINK, the embedding progress of entity context leaves little room for further improvement. \lee{We will subdivide the entity mentions and make statistics on the accuracy of different categories to conduct a more deeply exploration in our further work.}

We also expect that few-shot entity linking will rapidly develop and attract more enriching features like global coherence across all entities in a document, NIL prediction, joining entity retrieve and entity linking steps together, or providing complete end-to-end solutions.The latter would be an incredibly challenging task but also a fascinating research direction. Furthermore, performing named entity recognition and entity linking jointly to make these two tasks reinforce each other is a promising direction, especially for text where the named entity recognition tools perform poorly. \lee{Finally, we will explore to improve the quality of generated mention via leveraging the gap between the source and target domain.}

Our long-term vision is to design a domain-specific entity linking model to discover more entities and support various downstream applications. 

%% file: sec9-conclusion.tex
\section{Conclusion}\label{sec-conclusion}
In this paper, we conduct a comprehensive study of few-shot entity linking techniques, which is inevitable and challenging in the real world. 
To obtain more knowledge to deal with the ever more severe lack of labeled data, we propose a weak supervision strategy to generate in-domain synthetic data.
To further differentiate the quality of each synthetic data instance for model training, we design a meta-learning mechanism that can automatically assign different weights to the synthetic data.
We conduct comprehensive experiments on real-world datasets with reasonable measurements, and the results show our approach can significantly improve the state-of-the-art methods.

%% file: sec-ack.tex
\section*{Acknowledgment}
This work was supported in part by National Key Research and Development Program of China under Grant No. 2020YFA0804503, National Natural Science Foundation of China under Grant No. 61521002, Beijing Academy of Artificial Intelligence (BAAI), and Key Laboratory of Advanced Theory and Application in Statistics and Data Science, Ministry of Education. We further thank all anonymous reviewers for their constructive comments, which help us to improve the quality of this manuscript.

%% file: main.bbl
\begin{thebibliography}{10}
\providecommand{\url}[1]{#1}
\csname url@samestyle\endcsname
\providecommand{\newblock}{\relax}
\providecommand{\bibinfo}[2]{#2}
\providecommand{\BIBentrySTDinterwordspacing}{\spaceskip=0pt\relax}
\providecommand{\BIBentryALTinterwordstretchfactor}{4}
\providecommand{\BIBentryALTinterwordspacing}{\spaceskip=\fontdimen2\font plus
\BIBentryALTinterwordstretchfactor\fontdimen3\font minus
  \fontdimen4\font\relax}
\providecommand{\BIBforeignlanguage}[2]{{%
\expandafter\ifx\csname l@#1\endcsname\relax
\typeout{** WARNING: IEEEtran.bst: No hyphenation pattern has been}%
\typeout{** loaded for the language `#1'. Using the pattern for}%
\typeout{** the default language instead.}%
\else
\language=\csname l@#1\endcsname
\fi
#2}}
\providecommand{\BIBdecl}{\relax}
\BIBdecl

\bibitem{Auer2007DBpedia}
S.~Auer, C.~Bizer, G.~Kobilarov, J.~Lehmann, R.~Cyganiak, and Z.~Ives,
  ``Dbpedia: A nucleus for a web of open data,'' in \emph{The Semantic Web},
  K.~Aberer, K.-S. Choi, N.~Noy, D.~Allemang, K.-I. Lee, L.~Nixon, J.~Golbeck,
  P.~Mika, D.~Maynard, R.~Mizoguchi, G.~Schreiber, and P.~Cudr{\'e}-Mauroux,
  Eds.\hskip 1em plus 0.5em minus 0.4em\relax Berlin, Heidelberg: Springer
  Berlin Heidelberg, 2007, pp. 722--735.

\bibitem{fabian2007yago}
M.~Fabian, K.~Gjergji, W.~Gerhard \emph{et~al.}, ``Yago: A core of semantic
  knowledge unifying wordnet and wikipedia,'' in \emph{16th International World
  Wide Web Conference, WWW}, 2007, pp. 697--706.

\bibitem{bollacker2008freebase}
K.~Bollacker, C.~Evans, P.~Paritosh, T.~Sturge, and J.~Taylor, ``Freebase: a
  collaboratively created graph database for structuring human knowledge,'' in
  \emph{Proceedings of the 2008 ACM SIGMOD international conference on
  Management of data}, 2008, pp. 1247--1250.

\bibitem{miller1998wordnet}
G.~A. Miller, \emph{WordNet: An electronic lexical database}.\hskip 1em plus
  0.5em minus 0.4em\relax MIT press, 1998.

\bibitem{survey_vldbj}
X.~Qin, Y.~Luo, N.~Tang, and G.~Li, ``Making data visualization more efficient
  and effective: a survey,'' \emph{{VLDB} J.}, vol.~29, no.~1, pp. 93--117,
  2020.

\bibitem{deepeye_icde}
\BIBentryALTinterwordspacing
Y.~Luo, X.~Qin, N.~Tang, and G.~Li, ``Deepeye: Towards automatic data
  visualization,'' in \emph{34th {IEEE} International Conference on Data
  Engineering, {ICDE} 2018, Paris, France, April 16-19, 2018}.\hskip 1em plus
  0.5em minus 0.4em\relax {IEEE} Computer Society, 2018, pp. 101--112.
  [Online]. Available: \url{https://doi.org/10.1109/ICDE.2018.00019}
\BIBentrySTDinterwordspacing

\bibitem{mann2003unsupervised}
G.~Mann and D.~Yarowsky, ``Unsupervised personal name disambiguation,'' in
  \emph{Proceedings of the seventh conference on Natural language learning at
  HLT-NAACL 2003}, 2003, pp. 33--40.

\bibitem{artiles2007semeval}
J.~Artiles, J.~Gonzalo, and S.~Sekine, ``The semeval-2007 weps evaluation:
  Establishing a benchmark for the web people search task,'' in
  \emph{Proceedings of the fourth international workshop on semantic
  evaluations (semeval-2007)}, 2007, pp. 64--69.

\bibitem{yin2007object}
X.~Yin, J.~Han, and S.~Y. Philip, ``Object distinction: Distinguishing objects
  with identical names,'' in \emph{2007 IEEE 23rd International Conference on
  Data Engineering}.\hskip 1em plus 0.5em minus 0.4em\relax IEEE, 2007, pp.
  1242--1246.

\bibitem{han2009named}
X.~Han and J.~Zhao, ``Named entity disambiguation by leveraging wikipedia
  semantic knowledge,'' in \emph{Proceedings of the 18th ACM conference on
  Information and knowledge management}, 2009, pp. 215--224.

\bibitem{hachey2013evaluating}
B.~Hachey, W.~Radford, J.~Nothman, M.~Honnibal, and J.~R. Curran, ``Evaluating
  entity linking with wikipedia,'' \emph{Artificial intelligence}, vol. 194,
  pp. 130--150, 2013.

\bibitem{nvbench}
\BIBentryALTinterwordspacing
Y.~Luo, N.~Tang, G.~Li, C.~Chai, W.~Li, and X.~Qin, ``Synthesizing natural
  language to visualization {(NL2VIS)} benchmarks from {NL2SQL} benchmarks,''
  in \emph{{SIGMOD} '21: International Conference on Management of Data, China,
  June 20-25, 2021}, G.~Li, Z.~Li, S.~Idreos, and D.~Srivastava, Eds.\hskip 1em
  plus 0.5em minus 0.4em\relax {ACM}, 2021, pp. 1235--1247. [Online].
  Available: \url{https://doi.org/10.1145/3448016.3457261}
\BIBentrySTDinterwordspacing

\bibitem{dong2014knowledge}
X.~Dong, E.~Gabrilovich, G.~Heitz, W.~Horn, N.~Lao, K.~Murphy, T.~Strohmann,
  S.~Sun, and W.~Zhang, ``Knowledge vault: A web-scale approach to
  probabilistic knowledge fusion,'' in \emph{Proceedings of the 20th ACM SIGKDD
  international conference on Knowledge discovery and data mining}, 2014, pp.
  601--610.

\bibitem{lehmann2015dbpedia}
J.~Lehmann, R.~Isele, M.~Jakob, A.~Jentzsch, D.~Kontokostas, P.~N. Mendes,
  S.~Hellmann, M.~Morsey, P.~Van~Kleef, S.~Auer \emph{et~al.}, ``Dbpedia--a
  large-scale, multilingual knowledge base extracted from wikipedia,''
  \emph{Semantic web}, vol.~6, no.~2, pp. 167--195, 2015.

\bibitem{shi2018open}
B.~Shi and T.~Weninger, ``Open-world knowledge graph completion,'' in
  \emph{Thirty-Second AAAI Conference on Artificial Intelligence}, 2018.

\bibitem{diefenbach2018core}
D.~Diefenbach, V.~Lopez, K.~Singh, and P.~Maret, ``Core techniques of question
  answering systems over knowledge bases: a survey,'' \emph{Knowledge and
  Information systems}, vol.~55, no.~3, pp. 529--569, 2018.

\bibitem{yin2016simple}
W.~Yin, M.~Yu, B.~Xiang, B.~Zhou, and H.~Sch{\"u}tze, ``Simple question
  answering by attentive convolutional neural network,'' \emph{arXiv preprint
  arXiv:1606.03391}, 2016.

\bibitem{veyseh2016cross}
A.~P.~B. Veyseh, ``Cross-lingual question answering using common semantic
  space,'' in \emph{Proceedings of TextGraphs-10: the workshop on graph-based
  methods for natural language processing}, 2016, pp. 15--19.

\bibitem{hasibi2016exploiting}
F.~Hasibi, K.~Balog, and S.~E. Bratsberg, ``Exploiting entity linking in
  queries for entity retrieval,'' in \emph{Proceedings of the 2016 acm
  international conference on the theory of information retrieval}, 2016, pp.
  209--218.

\bibitem{corcoglioniti2016knowledge}
F.~Corcoglioniti, M.~Dragoni, M.~Rospocher, and A.~P. Aprosio, ``Knowledge
  extraction for information retrieval,'' in \emph{European Semantic Web
  Conference}.\hskip 1em plus 0.5em minus 0.4em\relax Springer, 2016, pp.
  317--333.

\bibitem{hasibi2015entity}
F.~Hasibi, K.~Balog, and S.~E. Bratsberg, ``Entity linking in queries: Tasks
  and evaluation,'' in \emph{Proceedings of the 2015 international conference
  on the theory of information retrieval}, 2015, pp. 171--180.

\bibitem{he2013learning}
Z.~He, S.~Liu, M.~Li, M.~Zhou, L.~Zhang, and H.~Wang, ``Learning entity
  representation for entity disambiguation,'' in \emph{Proceedings of the 51st
  Annual Meeting of the Association for Computational Linguistics (Volume 2:
  Short Papers)}, 2013, pp. 30--34.

\bibitem{newman2018jointly}
D.~Newman-Griffis, A.~M. Lai, and E.~Fosler-Lussier, ``Jointly embedding
  entities and text with distant supervision,'' \emph{arXiv preprint
  arXiv:1807.03399}, 2018.

\bibitem{radhakrishnan2018elden}
P.~Radhakrishnan, P.~Talukdar, and V.~Varma, ``Elden: Improved entity linking
  using densified knowledge graphs,'' in \emph{Proceedings of the 2018
  Conference of the North American Chapter of the Association for Computational
  Linguistics: Human Language Technologies, Volume 1 (Long Papers)}, 2018, pp.
  1844--1853.

\bibitem{deepeye_tkde}
Y.~Luo, X.~Qin, C.~Chai, N.~Tang, G.~Li, and W.~Li, ``Steerable self-driving
  data visualization,'' \emph{IEEE Transactions on Knowledge and Data
  Engineering}, vol.~34, no.~1, pp. 475--490, 2022.

\bibitem{JunyoungChung2014Empirical}
J.~Chung, C.~Gulcehre, K.~Cho, and Y.~Bengio, ``\BIBforeignlanguage{English
  (US)}{Empirical evaluation of gated recurrent neural networks on sequence
  modeling},'' in \emph{\BIBforeignlanguage{English (US)}{NIPS 2014 Workshop on
  Deep Learning, December 2014}}, 2014.

\bibitem{young2018recent}
T.~Young, D.~Hazarika, S.~Poria, and E.~Cambria, ``Recent trends in deep
  learning based natural language processing,'' \emph{ieee Computational
  intelligenCe magazine}, vol.~13, no.~3, pp. 55--75, 2018.

\bibitem{ncnet}
Y.~Luo, N.~Tang, G.~Li, J.~Tang, C.~Chai, and X.~Qin, ``Natural language to
  visualization by neural machine translation,'' \emph{IEEE Transactions on
  Visualization and Computer Graphics}, vol.~28, no.~1, pp. 217--226, 2022.

\bibitem{chisholm-hachey-2015-entity}
\BIBentryALTinterwordspacing
A.~Chisholm and B.~Hachey, ``Entity disambiguation with web links,''
  \emph{Transactions of the Association for Computational Linguistics}, vol.~3,
  pp. 145--156, 2015. [Online]. Available:
  \url{https://aclanthology.org/Q15-1011}
\BIBentrySTDinterwordspacing

\bibitem{kraljevic2021multi}
Z.~Kraljevic, T.~Searle, A.~Shek, L.~Roguski, K.~Noor, D.~Bean, A.~Mascio,
  L.~Zhu, A.~A. Folarin, A.~Roberts \emph{et~al.}, ``Multi-domain clinical
  natural language processing with medcat: the medical concept annotation
  toolkit,'' \emph{Artificial Intelligence in Medicine}, vol. 117, p. 102083,
  2021.

\bibitem{ratinov2011local}
L.~Ratinov, D.~Roth, D.~Downey, and M.~Anderson, ``Local and global algorithms
  for disambiguation to wikipedia,'' in \emph{Proceedings of the 49th annual
  meeting of the association for computational linguistics: Human language
  technologies}, 2011, pp. 1375--1384.

\bibitem{DBLP:conf/acl/LogeswaranCLTDL19}
\BIBentryALTinterwordspacing
L.~Logeswaran, M.~Chang, K.~Lee, K.~Toutanova, J.~Devlin, and H.~Lee,
  ``Zero-shot entity linking by reading entity descriptions,'' in
  \emph{Proceedings of the 57th Conference of the Association for Computational
  Linguistics, {ACL} 2019, Florence, Italy, July 28- August 2, 2019, Volume 1:
  Long Papers}, A.~Korhonen, D.~R. Traum, and L.~M{\`{a}}rquez, Eds.\hskip 1em
  plus 0.5em minus 0.4em\relax Association for Computational Linguistics, 2019,
  pp. 3449--3460. [Online]. Available:
  \url{https://doi.org/10.18653/v1/p19-1335}
\BIBentrySTDinterwordspacing

\bibitem{milne2008learning}
D.~Milne and I.~H. Witten, ``Learning to link with wikipedia,'' in
  \emph{Proceedings of the 17th ACM conference on Information and knowledge
  management}, 2008, pp. 509--518.

\bibitem{fevry2020empirical}
T.~F{\'e}vry, N.~FitzGerald, L.~B. Soares, and T.~Kwiatkowski, ``Empirical
  evaluation of pretraining strategies for supervised entity linking,''
  \emph{arXiv preprint arXiv:2005.14253}, 2020.

\bibitem{spitkovsky2012cross}
V.~I. Spitkovsky and A.~X. Chang, ``A cross-lingual dictionary for english
  wikipedia concepts,'' 2012.

\bibitem{cucerzan2007large}
S.~Cucerzan, ``Large-scale named entity disambiguation based on wikipedia
  data,'' in \emph{Proceedings of the 2007 joint conference on empirical
  methods in natural language processing and computational natural language
  learning (EMNLP-CoNLL)}, 2007, pp. 708--716.

\bibitem{gillick-etal-2019-learning}
\BIBentryALTinterwordspacing
D.~Gillick, S.~Kulkarni, L.~Lansing, A.~Presta, J.~Baldridge, E.~Ie, and
  D.~Garcia-Olano, ``Learning dense representations for entity retrieval,'' in
  \emph{Proceedings of the 23rd Conference on Computational Natural Language
  Learning (CoNLL)}.\hskip 1em plus 0.5em minus 0.4em\relax Hong Kong, China:
  Association for Computational Linguistics, Nov. 2019, pp. 528--537. [Online].
  Available: \url{https://aclanthology.org/K19-1049}
\BIBentrySTDinterwordspacing

\bibitem{wu-etal-2020-scalable}
\BIBentryALTinterwordspacing
L.~Wu, F.~Petroni, M.~Josifoski, S.~Riedel, and L.~Zettlemoyer, ``Scalable
  zero-shot entity linking with dense entity retrieval,'' in \emph{Proceedings
  of the 2020 Conference on Empirical Methods in Natural Language Processing
  (EMNLP)}.\hskip 1em plus 0.5em minus 0.4em\relax Online: Association for
  Computational Linguistics, Nov. 2020, pp. 6397--6407. [Online]. Available:
  \url{https://aclanthology.org/2020.emnlp-main.519}
\BIBentrySTDinterwordspacing

\bibitem{ganea-hofmann-2017-deep}
\BIBentryALTinterwordspacing
O.-E. Ganea and T.~Hofmann, ``Deep joint entity disambiguation with local
  neural attention,'' in \emph{Proceedings of the 2017 Conference on Empirical
  Methods in Natural Language Processing}.\hskip 1em plus 0.5em minus
  0.4em\relax Copenhagen, Denmark: Association for Computational Linguistics,
  Sep. 2017, pp. 2619--2629. [Online]. Available:
  \url{https://aclanthology.org/D17-1277}
\BIBentrySTDinterwordspacing

\bibitem{eshel-etal-2017-named}
\BIBentryALTinterwordspacing
Y.~Eshel, N.~Cohen, K.~Radinsky, S.~Markovitch, I.~Yamada, and O.~Levy, ``Named
  entity disambiguation for noisy text,'' in \emph{Proceedings of the 21st
  Conference on Computational Natural Language Learning ({C}o{NLL}
  2017)}.\hskip 1em plus 0.5em minus 0.4em\relax Vancouver, Canada: Association
  for Computational Linguistics, Aug. 2017, pp. 58--68. [Online]. Available:
  \url{https://aclanthology.org/K17-1008}
\BIBentrySTDinterwordspacing

\bibitem{peters-etal-2018-deep}
\BIBentryALTinterwordspacing
M.~E. Peters, M.~Neumann, M.~Iyyer, M.~Gardner, C.~Clark, K.~Lee, and
  L.~Zettlemoyer, ``Deep contextualized word representations,'' in
  \emph{Proceedings of the 2018 Conference of the North {A}merican Chapter of
  the Association for Computational Linguistics: Human Language Technologies,
  Volume 1 (Long Papers)}.\hskip 1em plus 0.5em minus 0.4em\relax New Orleans,
  Louisiana: Association for Computational Linguistics, Jun. 2018, pp.
  2227--2237. [Online]. Available: \url{https://aclanthology.org/N18-1202}
\BIBentrySTDinterwordspacing

\bibitem{devlin-etal-2019-bert}
\BIBentryALTinterwordspacing
J.~Devlin, M.-W. Chang, K.~Lee, and K.~Toutanova, ``{BERT}: Pre-training of
  deep bidirectional transformers for language understanding,'' in
  \emph{Proceedings of the 2019 Conference of the North {A}merican Chapter of
  the Association for Computational Linguistics: Human Language Technologies,
  Volume 1 (Long and Short Papers)}.\hskip 1em plus 0.5em minus 0.4em\relax
  Minneapolis, Minnesota: Association for Computational Linguistics, Jun. 2019,
  pp. 4171--4186. [Online]. Available: \url{https://aclanthology.org/N19-1423}
\BIBentrySTDinterwordspacing

\bibitem{zhou2018brief}
Z.-H. Zhou, ``A brief introduction to weakly supervised learning,''
  \emph{National science review}, vol.~5, no.~1, pp. 44--53, 2018.

\bibitem{Hospedales2021meta}
T.~M. Hospedales, A.~Antoniou, P.~Micaelli, and A.~J. Storkey, ``Meta-learning
  in neural networks: A survey,'' \emph{IEEE Transactions on Pattern Analysis
  and Machine Intelligence}, pp. 1--1, 2021.

\bibitem{DBLP:journals/jmlr/RaffelSRLNMZLL20}
\BIBentryALTinterwordspacing
C.~Raffel, N.~Shazeer, A.~Roberts, K.~Lee, S.~Narang, M.~Matena, Y.~Zhou,
  W.~Li, and P.~J. Liu, ``Exploring the limits of transfer learning with a
  unified text-to-text transformer,'' \emph{J. Mach. Learn. Res.}, vol.~21, pp.
  140:1--140:67, 2020. [Online]. Available:
  \url{http://jmlr.org/papers/v21/20-074.html}
\BIBentrySTDinterwordspacing

\bibitem{le2019distant}
P.~Le and I.~Titov, ``Distant learning for entity linking with automatic noise
  detection,'' \emph{arXiv preprint arXiv:1905.07189}, 2019.

\bibitem{DBLP:conf/icml/RenZYU18}
\BIBentryALTinterwordspacing
M.~Ren, W.~Zeng, B.~Yang, and R.~Urtasun, ``Learning to reweight examples for
  robust deep learning,'' in \emph{Proceedings of the 35th International
  Conference on Machine Learning, {ICML} 2018, Stockholmsm{\"{a}}ssan,
  Stockholm, Sweden, July 10-15, 2018}, ser. Proceedings of Machine Learning
  Research, J.~G. Dy and A.~Krause, Eds., vol.~80.\hskip 1em plus 0.5em minus
  0.4em\relax {PMLR}, 2018, pp. 4331--4340. [Online]. Available:
  \url{http://proceedings.mlr.press/v80/ren18a.html}
\BIBentrySTDinterwordspacing

\bibitem{sun2020meta}
S.~Sun, Y.~Qian, Z.~Liu, C.~Xiong, K.~Zhang, J.~Bao, Z.~Liu, and P.~Bennett,
  ``Meta adaptive neural ranking with contrastive synthetic supervision,''
  \emph{arXiv preprint arXiv:2012.14862}, 2020.

\bibitem{riedel2010modeling}
S.~Riedel, L.~Yao, and A.~McCallum, ``Modeling relations and their mentions
  without labeled text,'' in \emph{Joint European Conference on Machine
  Learning and Knowledge Discovery in Databases}.\hskip 1em plus 0.5em minus
  0.4em\relax Springer, 2010, pp. 148--163.

\bibitem{raiman2018deeptype}
J.~Raiman and O.~Raiman, ``Deeptype: multilingual entity linking by neural type
  system evolution,'' in \emph{Proceedings of the AAAI Conference on Artificial
  Intelligence}, vol.~32, no.~1, 2018.

\bibitem{onoe2020fine}
Y.~Onoe and G.~Durrett, ``Fine-grained entity typing for domain independent
  entity linking,'' in \emph{Proceedings of the AAAI Conference on Artificial
  Intelligence}, vol.~34, no.~05, 2020, pp. 8576--8583.

\bibitem{khalife-vazirgiannis-2019-scalable}
\BIBentryALTinterwordspacing
S.~Khalife and M.~Vazirgiannis, ``Scalable graph-based method for individual
  named entity identification,'' in \emph{Proceedings of the Thirteenth
  Workshop on Graph-Based Methods for Natural Language Processing
  (TextGraphs-13)}.\hskip 1em plus 0.5em minus 0.4em\relax Hong Kong:
  Association for Computational Linguistics, Nov. 2019, pp. 17--25. [Online].
  Available: \url{https://aclanthology.org/D19-5303}
\BIBentrySTDinterwordspacing

\end{thebibliography}
